# Hate Speech Classification Using SVM and Naive BAYES

## Asogwa D.C[1], Chukwuneke C.I[2], Ngene C.C[3], Anigbogu G.N[4]

*Department of Computer Science, Faculty of Physical Sciences, Nnamdi Azikiwe University,Awka. Anambra state Nigeria [1, 2, 3, 4].*
*dc.asogwa@unizik.edu.ng, ci.chukwuneke@unizik.edu.ng, cc.ngene@unizik.edu.ng,*
*gn.anigbogu@unizik.edu.ng*

*Abstract*
*The spread of hatred that was formerly limited to verbal communications has rapidly moved over the Internet. Social media and community forums that allow people to discuss and express their opinions are becoming platforms for the spreading of hate messages. Many countries have developed laws to avoid online hate speech. They hold the companies that run the social media responsible for their failure to eliminate hate speech. But as online content continues to grow, so does the spread of hate speech However, manual analysis of hate speech on online platforms is infeasible due to the huge amount of data as it is expensive and time consuming. Thus, it is important to automatically process the online user contents to detect and remove hate speech from online media. Many recent approaches suffer from interpretability problem which means that it can be difficult to understand why the systems make the decisions they do. Through this work, some solutions for the problem of automatic detection of hate messages were proposed using Support Vector Machine (SVM) and Naïve Bayes algorithms. This achieved near state-of-the-art performance while being simpler and producing more easily interpretable decisions than other methods. Empirical evaluation of this technique has resulted in a classification accuracy of approximately 99% and 50% for SVM and NB respectively over the test set.*
*Keywords: classification; hate speech; feature extraction, algorithm, supervised learning*

---------------------------------------------------------------------------------------------------------------------------------------

---------------------------------------------------------------------------------------------------------------------------------------

## I. INTRODUCTION:

Hate speech is an open speech that expresses hate or involves hostility towards a person or group based on something such as race, religion, sex, or sexual orientation. Hate speech is usually thought to include communications of animosity or disparagement of an individual or a group on account of a group characteristic such as race, colour, national origin, sex, disability, religion, or sexual orientation (*Brown-Sica et al, 2008*). One of the unique challenges of classifying hate speech is how to identify high level of rude, offensive or sometimes hateful language about the speech.

The task of classifying hate speech has been tested using a variety of labels: from offensive to non-offensives, abusive to non-abusive, harm or spam and so on. It has been observed that there is a big work surrounding text analysis of hate speech and similar topics such as rumors or spam. (Kumar S & Shah N, 2018).

The concept of social media speech interaction, such as Facebook and Twitter is a double-edged sword (Erlich Aaron et al, 2018). On the one hand, low cost and easy access to information and dissemination quickly push people to search for speech and know what is happening at the beginning of events with details and updates at the moment unlike newspapers or magazines in the olden days. On the other hand, it enables the widespread of hate speech because of its accessibility and lack of cost and control of the Internet. Reports indicate that the human ability to detect deception without special assistance is only 54%. So we need to use machine learning technique for classifying texts automatically (Krishnamurthy, Gangeshwar, et al, 2018).

Hate speech can be expressed in different forms. Explicit hate speech contains offensive words such as '*fuck*', '*asshole*'. Implicit hate speech can be realized by a sarcasm and irony (Waseem *et al*., 2017). While explicit hate speech can be identified using the lexicons that forms the hate speech, implicit hate speech is often hard to identify and requires semantic analysis of the sentence.

Hate speech classification is the prediction of the chances of a particular speech article (report, editorial, expose, etc.) being intentionally deceptive (Rubin, Conroy & Chen, 2015). The social media as well as other online platforms are playing an extensive role in the breeding and spread of hateful content eventually which leads to hate crime. In real world scenario, early classification of hate speech is very desirable to restrict the dissemination scope of hate speech and prevent the future propagation on social media. The proposed system will focus on development of machine learning model for hate speech classification using support vector machine (SVM) and naïve Bayes.

---

Support Vector Machine (SVM) is a supervised training model. This model projects the input feature vector into non-linear higher dimensional space. Then a linear decision boundary that maximizes minimum separation between training instances is constructed. SVM has been one of the classification algorithms used in many NLP tasks (Salminen *et al.*, 2020).

## II. RELATED WORKS:

Peter Burnap et al. (2016) employed a dictionary-based approach to identify cyber hate on Twitter. In this research, they employed an N-gram feature engineering technique to generate the numeric vectors from the predefined dictionary of hateful words. The authors fed the generated numeric vector to Machine Learning (ML) classifier, SVM and obtained a maximum of 67% F-score. Stéphan Tulkens et al. (2016) used a dictionary-based approach for the automatic detection of racism in Dutch Social Media. In this study, the authors used the distribution of words over three dictionaries as features. They fed the generated features to the SVM classifier. Their experimental results obtained 0.46 F-Score. Njagi Dennis et al. (2015) used ML-based classifier to classify hate speech in web forums and blogs. The authors employed a dictionary-based approach to generate a master feature vector. The features were based on sentiment expressions using semantic and subjectivity features with an orientation to hate speech. Afterward, the authors fed the masters feature vector to a rule-based classifier. In the experimental settings, the authors evaluated their classifier by using a precision performance metric and obtained 73% precision.

Nonetheless, the combination of dictionary-based and ML approaches showed a good result. However, the major disadvantage of such type of approach is that it requires a dictionary, based on the large corpus to look for domain words. To overcome this drawback, many of the researchers have used a bag of words (BOW) based approach which is similar to a dictionary-based approach but the word features are obtained from training data and not from the predefined dictionaries.

Irene Kwok et al. (2013) employed an ML-based approach to the automatic detection of racism against black in the twitter community. In their research, they employed unigram with the BOW-based technique to generate the numeric vectors. The authors fed the generated numeric vector to the Naïve Bayes classifier. Their experimental results obtained a maximum of 76% accuracy. Sharma et al. (2015) classified hate speech on twitter. In their research, they employed BOW features. The authors fed the generated numeric vector to the Naïve Bayes classifier. Their experimental results showed a maximum of 73% accuracy.

Nevertheless, BOW showed better accuracy in social network text classification. However, the major disadvantage of this technique is, the word-order is ignored and causes misclassification as different words are used in different contexts. To overcome this limitation, researchers have proposed an N-grams-based approach (Cavnar, W.B. & J.M. Trenkle, 1994)

Zeerak Waseem et al. (2016) classified hate speech on twitter. In their research, they employed character Ngrams feature engineering techniques to generate the numeric vectors. The authors fed the generated numeric vector to the LR classifier and obtained overall 73% F-score. Chikashi Nobata et al. (2016) used the ML-based approach to detect the abusive language in online user content. In their research authors employed character Ngrams feature representation technique to represent the features. The authors fed the features to the SVM classifier. The results showed that the classifier obtained overall 77% F-score. Shervin Malmasi et al (2017) used an ML-based approach to classify hate speech in social media. In their research, the authors employed 4grams with character grams feature engineering techniques to generate numeric features. The authors fed the generated numeric features to the SVM classifier. The authors reported maximum of 78% accuracy.

The N-gram-based approach gives better results than the BOW-based approach but it has two major limitations. First, the related words may be at a high distance in sentence and finally increasing the N value, resulting in slow processing speed, proposed by Chen, Y., (2011)

In recent years, authors employed deep learning-based NLP techniques to classify hate speech messages. Köffer et al. (2018) employed word2vec features and SVM classifiers to classify German texts hate speech messages and obtained 67% F-score. The word2Vec showed the lowest results because such approaches need enormous data to learn complex word semantics.

There has been a good attempt to construct and detect hate speech as well as offensive language in other languages (i.e. Danish). An important research study of Sigurbergsson G. I., & Derczynski, L., (2019) worked on the construction of Danish dataset for hate speech and offensive language detection. The dataset contained comments from Reddit and Facebook. It also contained the various types and targets of the offensive language. The authors achieved the highest F1 score of 0.74 by using deep learning models with different features sets.

Tian et al, 2021 discussed NLP and its utilities, as well as commonly employed features and classification methods in hate speech detection. They also explored the challenges in the field of hate speech detection and emphasized on the importance of standardized methodologies for building corpora and data sets.

Schmidt et al. (2017) conducted a survey on hate speech detection using natural language processing. The authors discussed in detail studies regarding various feature engineering techniques to be used for supervised classification of hate speech messages. The major drawback of this survey is that there were no experimental results for those mentioned techniques.

Sindhu A et al (2020), proposed the study of employed automated text classification techniques to detect hate speech messages. Moreover, the study compared three feature engineering techniques and eight ML algorithms to classify hate speech messages. The experimental results exhibited that the bigram features, when represented through TFIDF, showed better performance as compared to word2Vec and Doc2Vec features engineering techniques. Moreover, SVM and RF algorithms showed better results compared to LR, NB, KNN, DT, AdaBoost, and MLP. The lowest performance was observed in KNN. The work has two important limitations. First, the proposed ML model is inefficient in terms of real-time predictions accuracy for the data. Finally, it only classifies the hate speech message in three different classes and is not capable enough to identify the severity of the message. Previous studies showed that a variety of researchers from across the globe are working on hate speech recognition written in different languages such as German, Dutch and English.

Hate speech classification/detection has received a lot of attention recently due to their impact on users' security.

## III. MATERIALS AND METHODS:

The CRISP-DM (cross industry standard process for data mining) was used in this research because it comprises of steps needed to achieve the goal of this work. It is a comprehensive data mining methodology and process model that provides anyone with a complete blueprint for conducting a data mining project. Supervised learning algorithms, support vector machine and Naïve Bayes were used to implement the hate speech detection/classification using WEKA machine learning tools, java programing language, NetBeans IDE.

The following steps were followed to achieve the desired result:
1)      Collection of the sample data based on Learning classification
2)      Preprocessing (that is the dataset was provided with labels), since it is a supervised learning approach.
3)      Applying feature extraction with Weka library (to convert the attribute into a numeric values for classification analysis)
4)      Resampling the dataset by applying training set and testing set during system development analysis using Weka tools.
5)      Data Cleaning: The data cleaning task in this research is focused on noisy data removal. The cleaning process solves the duplicated records problem. The size of data sets is very big; it needs huge manual effort for pre-processing. To accomplish this task manually is neither quick nor precise, since the human error can occur. As a result, the research designed and implemented a pre-processing software tool using Java programming language to take the advantage of automated pre-processing methods and save data miner's time. After applying Weka tool on the received data set. The sizes of the collected data were reduced from the instances.
6)      Data Transformation: The data sets in its real form are difficult to be applied, so some transformation techniques were required to get the appropriate form such as the normalization and nominal to numeric conversion, etc. Data transformation encompasses into two sub-tasks, data normalization, and data conversion.
i.)  Data Normalization: The algorithm will be able to give a better result, reduce calculation time, and speeds learning process when the data values  are small and fall into a specific range that can be either [-1, 1] or [0, 1] depending on data analyst's choice.
ii.)  Data Sets Conversion: Data conversion means converting data values from nominal values to numeric values. Generally, the Naïve Bayes (NB) and Support vector machine which was adopted for the hate speech analysis tends to perform worst when it deals with nominal values. Therefore, replacing nominal test results with numeric ones was one of the preprocessing steps.
7)      Data Reduction:  This step is a part of data preprocessing, it focuses on attribute selection, where relevant attributes were selected, whereas irrelevant attributes were removed from the data set. The potential benefits of feature selection are improving the prediction performance of the predictors, providing faster, more cost effective predictors, and reducing training time. Figure 3.0 shows the different steps followed to achieve the desired result.

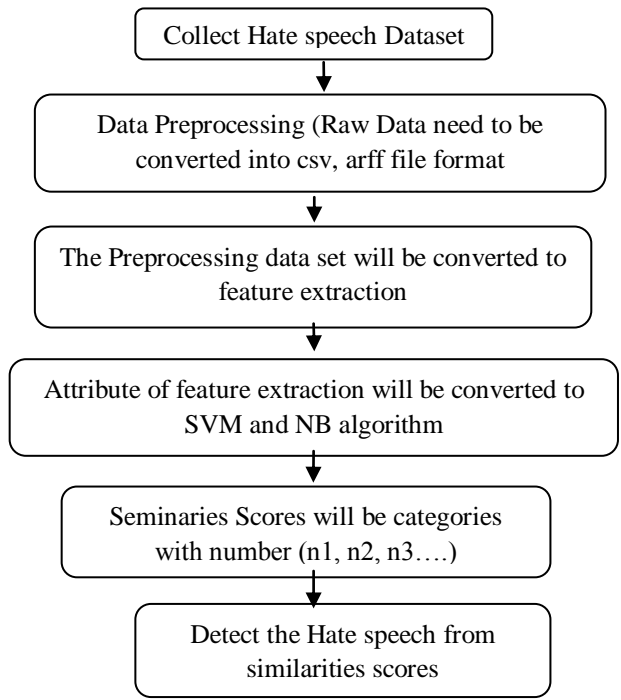

**Figure 3.0: processing steps**

**3.1: Algorithms used for the classification and detection of hate speech:**
**a) Support Vector Machine**

The Support Vector Machine (SVM) algorithm is one of the supervised machine learning algorithms that is employed for various classification problems. It has its applications in credit risk analysis, medical diagnosis, text categorization, and information extraction. SVMs are particularly suitable for high dimensional data. There are so many reasons supporting this claim. Specifically, the complexity of the classifiers depends on the number of support vectors instead of data dimensions, they produce the same hyper plane for repeated training sets, and they have better generalization abilities

**b)Naïve Bayes**

These are probabilistic classifiers commonly used in machine learning. However, the Bayesian classifiers are statistical and also possess learning ability. Multinomial model is used by Naïve Bayes for large datasets. The performance could be enhanced by searching the dependencies among attributes. It is mainly used in data pre-processing applications due to ease of computation. Bayesian reasoning and probability inference are employed in predicting the target class. Attributes play an important role in classification. Therefore the performance of Naïve Bayes depends on the accuracy of the estimated conditional probability terms. It is hard to accurately estimate these terms when the training data is scarce.

The Naïve Bayes classifier technique is based on Bayesian theorem, whereas it performs better when data dimensionality is high (Nikam 2015). The Bayesian classifier is capable of calculating the most possible output based on the input. There is no problem to add new raw data at run time and have a better probabilistic classifier.

# IV. SYSTEM DESIGN:

Data set collected from online resources was used to form the basis of the system analysis. The data collected passed through the preprocessing stage, feature set extraction, training set, validation set and testing set in other to remove noise (error) from the sample data collected via unstructured data set before it can be used for analysis. This sample was further applied to the classification techniques, SVM and Naïve Bayes in other to yield the promising results. Figure 4.0 depict the clear understanding of the system analysis of the work.

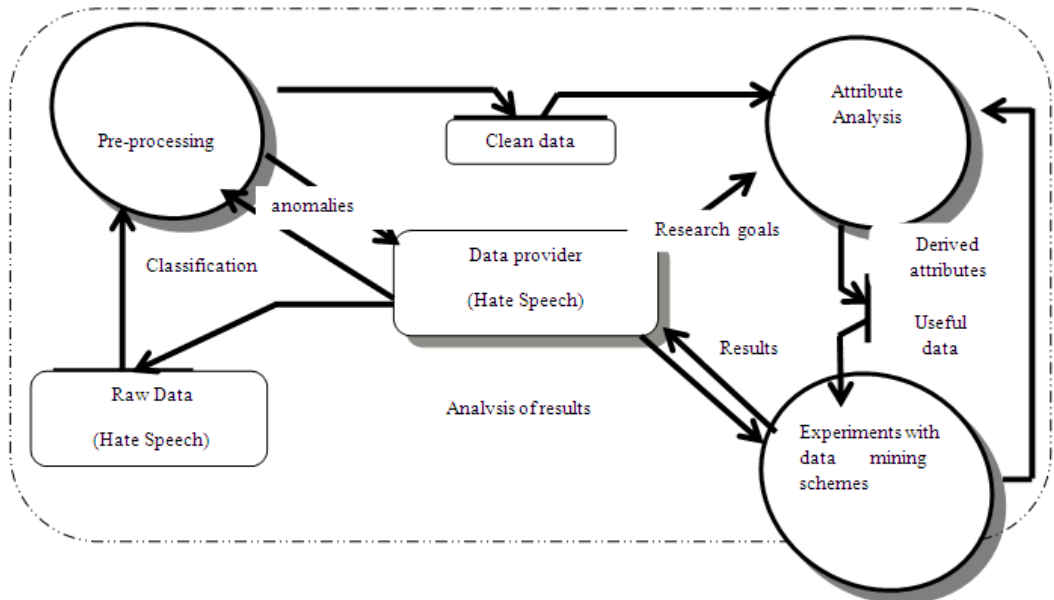

**Figure 4.0: the system design**

**4.1 System Implementation**:

All the experiments were carried out using open source machine learning tool Weka 3.9.1 and java programming language with Netbeans IDE and with machine learning classification. The work was implemented with the set of 17 feature set or attributes to distinguish their performance when those factors were structured into the Weka plugin and java Netbeans to classify those sample dataset which were implemented as an information table. The limited sample collected via online resource with (17) attributes and (62485) instances was used to perform the analysis, used to build the model for predicting a promising result. Here the numeric values from a given sample which was transformed into an excel format with an extension of csv and arff for machine readable task. Figure 4.1 shows the practical understanding of this work

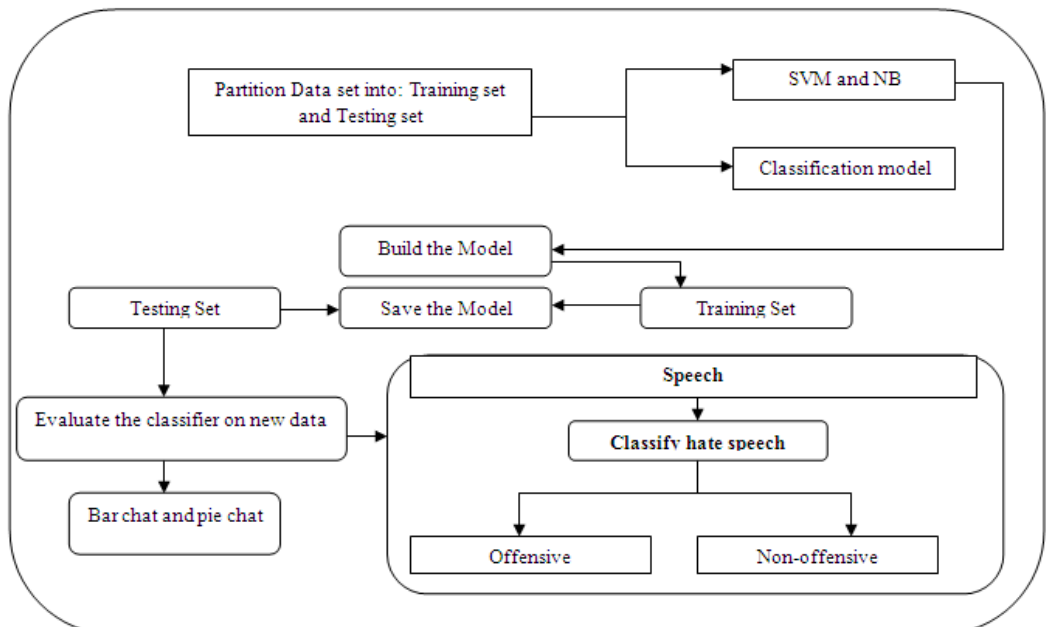

Figure 4.1: the system implementation

**4.2     Evaluation Matric**

A true positive (TP) and false positive (FP) in classification is the number of instances that are correctly classified. A true positive is the number of classifications that the classifier classified correctly as the answer. A true negative is the opposite and is the amount of classifications that the classifier correctly predicted not to be the answer. The true negative (TN) and false negatives (FN) are the incorrectly classified instances. A false positive happens when an incorrect instance is classified as correct. A false negative occurs when a correct instance is classified as incorrect.

## V.    RESULTS AND DISCUSSION:

The experiment of hate speech detection and classification was done in two folds; the sample of the dataset collected was used to perform Hate speech detection and classification and the training set was used to build the model. The test set was then used for predicting the result with unknown class label to predict a new class label with their respective classes. The two classes were labeled offensive and non-offensive speech.

### 5.1 Evaluation Of Experimental Result For Naïve Bayes
Model Information
================

| | | |
|---|---|---|
| Correctly Classified Instances | 9983 | 79.8832 % |
| Incorrectly Classified Instances | 2514 | 20.1168 % |
| Kappa statistic | 0.6013 | |
| K&B Relative Info Score | 643063.9319 % | |
| K&B Information Score | 6347.927 bits | 0.508 bits/instance |
| Class complexity \| order 0 | 12336.1197 bits | 0.9871 bits/instance |
| Class complexity \| scheme | 15728.4368 bits | 1.2586 bits/instance |
| Complexity improvement (Sf) | -3392.3171 bits | -0.2715 bits/instance |
| Mean absolute error | 0.2445 | |
| Root mean squared error | 0.3837 | |
| Relative absolute error | 49.7949 % | |
| Root relative squared error | 77.4254 % | |
| Total Number of Instances | 12497 | |

### 5.2a Details of classification

| | TP Rate | FP Rate | Precision | Recall | F-Measure | ROC Area | Class |
|---|---|---|---|---|---|---|---|
| | 0.884 | 0.266 | 0.717 | 0.884 | 0.792 | 0.921 | non-offensive |
| | 0.734 | 0.116 | 0.892 | 0.734 | 0.805 | 0.922 | offensive |
| Weighted Avg. | 0.799 | 0.181 | 0.816 | 0.799 | 0.799 | 0.922 | |

### 5.2b Confusion matrix
```
  a    b   <-- classified as
4788  627 |   a = non-offensive
1887 5195 |   b = offensive
```
**Model Accuracy** for Naïve Bayes is 50.0%

### 5.3 Evaluation Of Experimental Result For Svm
Model Information
================

| | | |
|---|---|---|
| Correctly Classified Instances | 12368 | 98.9678 % |
| Incorrectly Classified Instances | 129 | 1.0322 % |
| Kappa statistic | 0.979 | |
| K&B Relative Info Score | 1214007.2298 % | |
| K&B Information Score | 11983.9239 bits | 0.9589 bits/instance |
| Class complexity \| order 0 | 12336.1197 bits | 0.9871 bits/instance |
| Class complexity \| scheme | 22197.4947 bits | 1.7762 bits/instance |
| Complexity improvement (Sf) | -9861.375 bits | -0.7891 bits/instance |
| Mean absolute error | 0.0153 | |
| Root mean squared error | 0.0979 | |
| Relative absolute error | 3.1067 % | |
| Root relative squared error | 19.759 % | |
| Total Number of Instances | 12497 | |

### 5.3a Details by classification

| | TP Rate | FP Rate | Precision | Recall | F-Measure | ROC Area | Class |
|---|---|---|---|---|---|---|---|
| | 0.985 | 0.007 | 0.991 | 0.985 | 0.988 | 0.994 | non-offensive |
| | 0.993 | 0.015 | 0.988 | 0.993 | 0.991 | 0.994 | offensive |

Weighted Avg.    0.99    0.011    0.99    0.99    0.99    0.994

**5.3b Confusion matrix**

      a     b    <-- classified as
   5333    82 |    a = non-offensive
     47  7035 |    b = offensive

**Accuracy**: 99.37% Application model SVM

The result of this work was obtained using a set of Hate speech dataset on the model evaluations built from the training data set which was shown above based on the proposed model used to classify the total number of 12495 instances. The accuracy of the model for SVM is 99.37% and NB is 50.00%. That is, the SVM gave a higher accuracy when compared to NB.

## VI.    SUMMARY AND CONCLUSION

Recent study shows that, researchers have proposed various techniques to detect hate speech. Basically, major works done on dynamic hate speech detection are biased towards limited feature space. Therefore, there is a need to search for new features for hate speech detection. The behavior of hate speech in a digital attack were explored. This work was carried out to develop a system for Hate Speech detection and classification using Naïve Bayes and SVM. The program was written using java programming language and Weka tools as classifiers. The dataset used were collected from online resources called Unique Client Identifier (UCI) for machine learning repository system. It is  used for a collection of dataset, domain theories, and data generators which are used by the machine learning community for the empirical analysis of machine learning algorithms. Furthermore, the hyper-parameters for each classifier, the SVM and Naïve Bayes were tuned to achieve the best performance measures. It was observed that the regularized SVM achieved the best results compare to Naïve Bayes used for the experiment. The detection was based on two classes, offensive and non-offensive labels.

**6.1 Further Work**

This work was focused on hate speech detection and classification using machine learning algorithms, namely Naïve Bayes and SVM for extracting the features of the hate speech. Researchers in future should do a comparison of other algorithms like Random Forest, k-nearest neighbors' algorithm (k-NN), deep learning and Artificial Neural Network (ANN) to determine which will be most optimal in hate speech detection.